\documentclass[preprint,12pt]{ieeeconf}  
\usepackage{graphicx}                       
\graphicspath{ {image/}}
\usepackage{amssymb}
 
\usepackage{lipsum}
\usepackage{mathtools}
\usepackage{cuted}
\usepackage[utf8]{inputenc}
\usepackage{amsmath}
\usepackage{lipsum}
\usepackage{graphics}
\usepackage{authblk}
\usepackage{etoolbox}
\AtBeginEnvironment{tabular}{\tiny}

\begin{document}

\title{\LARGE \bf Concepts of Self-maintaining Robots and Their Design}
\author[1]{\bf{Chenjie SHI} }
\author[2]{\bf{Sandor M Veres} }
\affil[$ $]{Department of Automatic Control and Systems Engineering \\ University of Sheffield, S1 3JD, United Kingdom } 



\maketitle





\begin{abstract}
This paper proposes an initial theory for robotic systems that can be fully self-maintaining.  The new design principles focus on functional survival of the robots over long periods of time without human maintenance. Self-maintaining semi-autonomous mobile robots are in great demand in nuclear disposal sites from where their removal for maintenance is undesirable due to their radioactive contamination. Similar are requirements for robots in various  defence tasks or space missions.  For optimal design, modular solutions  are balanced against capabilities to replace smaller components in a robot by itself or by help from another robot. Modules are proposed for the basic platform, which enable self-maintenance within a team of robots helping each other.  The primary method of self-maintenance is replacement of malfunctioning modules or components by the robots themselves. Replacement necessitates a robot team's ability to diagnose and replace malfunctioning modules as needed. Due to their design, these robots still remain manually re-configurable if opportunity arises for human intervention. Apart from the basic principles, an evolutionary design approach is presented and a first mathematical theory of the reliability of a team of self-maintaining robots is introduced. 
\end{abstract}

\emph{Robot intelligence,  self maintenance, reliability theory}

\footnotetext[1]{{Email:cshi17@sheffield.ac.uk}}
\footnotetext[2]{{Email:s.veres@sheffield.ac.uk}}

\section{Introduction}
In the past decades, development and application of robots has served an increasing range of areas from room cleaning to space exploration. Robotic researchers and engineers developed more functionality, flexibility, adaptability and reliability of new robotic systems. Modular robotic systems, in particular, enjoyed great advances in the past two decades. 

However, following the wide-range application of robots, the maintenance of robots has also become a problem, incurring more expense, time and human resource for detection and repair. 
Lack of the ability to self-maintain is a crucial contributor to the  high long term cost of robotic systems. Past and recent research has addressed re-configurable and modular robotic systems and swarms. Self maintenance has, however, received little focused analysis, if any at all.

This paper aims to provide the first steps in the analysis, requirement formulations and reliability computations for self-maintaining robotic systems.

Modular robotic systems have been derived from dynamically re-configurable robotic system (DRRS) or cellular robotic systems by Fukuda \cite{c2}, the principles of which have been transferred to modular robotic systems.  DRRS took its inspiration from observing living creatures, where cells form  novel configurations to create new functionality when they assemble, although one cell itself may only have a simple structure and functionality. 

From 1998, reconfiguration for specific tasks began to dominate the modular robotic area with the utilisation of lattice-based structures. Qiao et al. \cite{c3} put forward a modular robotic design and implementation,  which is a single robot with three joints and can have the ability to do rectilinear motion, lateral shift, lateral rolling and rotation with a convenient "pin-hole-based" docking system. Moreover, Baca et al. \cite{c4} developed a homogeneous modular robotic system called ModRED, capable of self-reconfiguration and to transform into a snake-like robot with a genderless docking mechanism. 

Apart from homogeneous modular robotic systems, Baca et al. \cite{c5} proposed a heterogeneous modular robotic system named SMART, which includes three different types of modules, so called 'M-robots' and colonies. Compared with Qiao's work \cite{c3} and ModRED \cite{c4}, SMART \cite{c5} exhibits high flexibility and functionality. The paper \cite{c1} by  Ahmadzadeh et al. has shown that, compared with modular robotic systems, fixed-shapes or fixed functionality of robots  limit the performance in unpredictable environments and in a variety of tasks. From the inspiration gained from flexibility, adaptability and self-organizing properties of multi-cellular biological system,  a number of roboticists created self-organizing machines that adapt themselves to unexpected conditions. 

This paper is  specific about long term self-maintenance of a group of robots, which has not yet been addressed in prior publications to the best of our knowledge.  Section 2 introduces essential concepts,  Section 3 addresses hardware design, reliability computations for single and groups of robots are provided in Section 4, Section 5  considers the module redundancy allocation problem, Section 6 gives an overview on possible  universal module connectors, and finally conclusions are drawn.

\section{Concepts for  Self-maintaining Robots}

This section examines options for robot group design and their modules for the development of self-maintaining robots.

\subsection{SMR grouping types}


The problem of self maintenance of robots can be considered for 

(a) a single robot; 

(b) a homogeneous set of robots with similar architectures;

(c) a heterogeneous set of robots with varying architectures.    
	
\subsubsection{Single robot self maintenance}

In case of a single robot self-maintaining itself, it needs technical solutions, both in mechatronics and software, which would keep its performance high for long periods of time. These solutions can include  an architecture, which optimises the probability of the robot being capable of replacing failing components and able to reconfigure its software. 
Such an architecture would inevitably include distributed computing of actuators to handle components reliably and combined with some redundancy of the actuators in case they need to be replaced by the robot itself. Distributed software needs to be able to recover from any failure expect if all processors stop working at the same time, in which case the robot deemed to fail and this probability of total failure needs to be kept low. 

There is a lot to learn from solutions of aviation safety solutions developed for many years for passenger aircraft \cite{article} flying today, which achieved remarkably high reliability levels. Though the similarity is there during flights, the difference is that our robots are to self-maintain for long time periods, unlike aircraft checked and maintained by technical staff between flights. 
Some ideas of this article, in the use of both software and hardware redundancy, originate from aviation systems.

\subsubsection{Sets of robots}
	
Use of a team of robots, who can help each other, can significantly increase the probability that the team's performance can be maintained. Total failure of any single robot may be recoverable to full functionality. Sets of robots with homogeneous architecture, where each robot has the same architecture, can simplify the overall design and assessment of reliability. It also simplifies the skills set the robots need for repairs. 

Heterogeneous set of robots, which have different 
architectures, likely to face more challenging problems when aiding each other in case of faults. On the other hand, heterogeneous robots may be needed for effective work in some applications, hence they remain an important case for reliability assessment.

\subsection{Modules and components}

A component in a robotic system is a small part that can be  less expensive to replace than a module when it fails. For easy handling and replacement by the robots themselves, components should be plug-and-play (PAP). For instance, sensors and gripper actuators can be made self-testing and  PAP  replaceable and so can be mobility components such as wheels and motors.

Modules represent combinations of components, which together serve a well defined functionality and as such are simple to replace by PAP for mechanical, computing hardware and  software modules, and also for connectivity and communication modules. 

Design and maintenance can be simplified if robots are based on a modular-architecture rather than on a components-based architecture. Here we assume that a module is an interconnection of a set of components. On the other hand, component replacement can be less expensive and less wasteful, although the development of robot abilities to discover faults and replace components can come with higher overall system complexity.  

It is often difficult to decide whether a module should be called a component, or a component is to be called a module. For this reason, we unify these concepts in this paper and we refer to all PAP replaceable parts of a robot as "modules". Calculations of economic structural design and costs can decide, which parts of a robots design should then become modules. 

Based on the principles outlined above, the design requirements of self-maintaining robots (SMRs) can be introduced as follows.

\vspace{1mm}

\noindent \emph{Definition 3.1} 
	A single robot or a team of robots is called \emph{self maintaining at robustness level k} if it satisfies the following conditions.  If, during full functionality, not more than $k$ modules malfunction at the same time, then   
	\begin{enumerate}
		\item[(a)] they are able to identify all failing modules;
		\item[(b)] they are able to replace all failing modules.
	\end{enumerate}
	If, in addition, the time averaged costs of module  failures and replacements is minimised by design, then they are called  \emph{level  k  optimized self maintaining robots}. Section 5 considers the associated optimal redundancy allocation problem.
	
	It is much needed in practical robotic applications that 
	{\it long term average cost} of self-maintenance is assessed and quantified. The computations presented in Section 4 aim to help this assessment.  
		\vspace{1mm}
		
\subsection{System reliability}

Reliability of self maintenance  depends  on the following factors:

 \begin{enumerate}
 	\item Reliability of robot components under different environmental conditions. 
 	\item Redundancy of critical components on each robot.
 	\item Ability of self-reconfiguration by and of the software on each robot.
 	\item Communication and functional monitoring ability among robots.
 	\item Physical and communications skills of each robot to do repair work on other robots.  
 \end{enumerate}

Neither self-maintenance of a single robot nor that of a team of robots can be guaranteed with probability one. As with civil aircraft, which needs to be self-sufficient during a flight, high probability of keeping up performance need to be guaranteed, but in this case without human maintenance. In aviation this is assessed by quality assurance of components and structures and by some redundancy of control devices and redundancy in software \cite{article}. 
Section 4 presents quantitative models of reliability for continuous operations of robot. 
	
	\vspace{2mm}


\section{Hardware Design for Self Maintenance}

By the nature of of robotic development, the theory of self maintenance needs to address two main aspects: hardware self-maintenance and software self-maintenance. 
Concerning the hardware, which includes mechanical components, actuators and the electronics of digital control and computing, 
this section proposes six types of {\it core modules} to start from. 
This organisation around core modules does not exclude the possibility that the core modules have one or more {\it sub-modules}, which are replaceable in a plug-and-play manner.

A most fundamental requirement for hardware self-maintenance is that the modules should have attributes such as easy 
diagnosis and replacement.
These principles can be aided by providing suitable mechanical designs of connectors for each hardware module, so that the manipulator can easily handle fitting. 
Consequently, the {\it manipulator module} is a core part of any SMR system. Apart from manipulators, locomotion  and  processor modules are  also vital on robots for their perception, reasoning, planning, commanding movements by control signals. 
Here we provide brief overviews for six types of core modules. 


\subsection{Platform/locomotion modules}

Platform modules are crucial in that they deliver mobility for a robot. 
With their driving mechanisms, platform modules can provide various locomotion methods such as on the ground and aerial motion. Apart from its power subsystem with batteries, a platform module needs to hold a docking system to enable connection with other modules. Docking can rely on electro-mechanical connectors, which attach various payload modules to it. A platform module will likely to have various  sub-modules for possible replacements.  

\subsection{Power modules and components }

Power modules need to contain  powerful batteries for energy supply and some sensors, which help processors to detect the available power and conditions of the battery. The latter is important to predict expected failure times. 

\subsection{Processor modules}

Processor modules are computational
units that process sensory signals and use them in the execution of   movements and manipulation tasks. Some processors can do perception and reasoning, planning and decision making for ongoing work as well as for self-maintenance. 
For hardware reliability of a robot, the processor architecture needs to be such that operations must be recoverable if some processors fail. 


\subsection{Communication modules}

Communication modules  can provide two kinds of capabilities. The first is to provide a network of communication between modules of a single robot 
and a second is to provide the means of communication of the reasoner processor module with other robots. For safety there can be redundant, secondary set of communication channels, both inside a robot as well as among robots. Among robots {WiFi} communication can be physically disturbed and alternative communications by sound, light, visual signalling and vibration are practical alternatives to provide reliability. 
Under some extremely challenging condition, such as in nuclear waste silos, the alternatives of communication technology may be critically important to deploy. Secure communication with remote human supervisors also necessitates the 
 use of alternative communications. 

\subsection{Manipulator modules and sub-modules}

Manipulator modules play a pivotal role in the operational life of self-maintaining robots. Most of the payload/functional modules, which serve the purpose rather than just the mere survival of a robot, may also heavily rely on the manipulator module. 
In most missions of a robot team, if there is no manipulator module left functioning, then the robotic system is likely to have lost its self-maintaining ability. 

\subsection{Modules for hazardous environments}

The practical areas where self-maintaining robots are needed are nuclear waste processing, long term defence tasks and space missions on space stations or moons, asteroids and planets of the solar system. These are hazardous environments.  In some of these, the detection and modelling of radiation levels in the robot's environment are of vital importance.  The radiation protection modules  
can be used to sense and compute spacial radiation models to inform the planning of the processor module and thereby protect the robot from avoidable harm. 
The module can also include a radiation shielding controller. Other economic areas, where self maintenance is useful, but less vital, are agriculture, food production and manufacturing of goods in the future.

\section{System reliability}

Reliability theory of SMR systems can be based on probabilistic models, which compute the likelihood of system breakdown during a given time period. Apart from their use in design, these models can also be applied in the SMR system's decision making in the interest of self-maintenance as described later. Here a  probabilistic model is presented  for the optimisation of redundancy allocation in an SMR system with an example of design optimisation offered by a genetic algorithm in Section 5.

\subsection{Problem description}
Following the above definitions, this subsection introduces a system reliability model and redundancy allocation criterion. 
The reliability problem needs to address two relevant aspects of self-maintenance: the first is structural reliability in terms of hardware redundancy. The second is functional reliability in terms of ability to reconfigure. 
All methods in this section contribute to structural reliability by redundancy allocation that includes choice of a component's reliability from their available range. 

\subsection{Redundancy Models}
Here a two-layer series-parallel  system  is considered for structural redundancy and its simplicity:
\begin{itemize}
  \item Robot team $\textit R_{gr}$: consists of a relevant group of robot subgroups within each with similar capabilities. 
  \item Single robot $R_j$: composed of modular structure with  redundancies of modules within each robot, 
\end{itemize}



At the robotic team level, a series of robot subgroups with similar capabilities, can provide shared redundancy. 
For a single robot, similar modules can provide redundancy of a series of functional capabilities. 

\subsubsection{Robot and module redundancies}

To increase reliability, most complex system can adopt redundancy
technology \cite{rausand2003system} . Redundancy techniques involve the application of  both robotic unit redundancy
and component redundancy. In our case component redundancy refers to redundancy of functional modules within a robot.

\subsubsection{Reliability deficiency/failure detection} 

In most cases switching means that after due to an internal request, a cold standby component replaces a formerly active component in the event of the failure of the active component. However, the possible failure of switching to a cold standby component also needs to be analyzed for its likelihood, 
which introduces the concept of switching reliability. And in SMR theory, switch reliability also covers the detection factor, which means the switch reliability can be regarded as a integrated parameter of switch reliability and detection reliability. Furthermore,  switching  has some inherent complexity due to possible module replacements across the groups of robots, not only within a single robot.  

 Assume that a robotic system $\textit R_{all}$ is composed of homogeneous robots $\textit R_{j}$, with $\textit m_{i}$ number of different modules. 
The reliability of switching of a module with index $i$ will be denoted by probability $p_i$. The associated prior probability of the detection of a module failure will be denoted by $p^d_i$. It is however a practical simplification to limit detection probability to the ability of the module to self diagnose, which clearly is an under-estimate of the true value but easier to determine. Complete module failure is not an obstacle of detection if the approach taken is to use liveliness signals for all modules, collected by all other modules onboard of a robot. Detection of a whole robot's failure can be detected by other robots due to that it stops broadcasting its liveliness signals. 

The switching over reliability of a component with index $i$  by a robot or by the whole team will be denoted by probabilities $p^s_i$ and $p^o_i$, respectively.  

\subsubsection{Redundancy capacity of robots}
Each robot has an upper-limit  $w_i$ for the number of modules it can hold in standby. These cold-standby modules, however, do not need to be physically carried by each robot. It is more power efficient if they can pick them up from a, possibly  
shared, module station. Storing two many spare modules can also be uneconomical in practice. 


\subsubsection{Cold standby redundancy and active redundancy}
In reliability studies, we find two mainstream strategies: cold standby and active redundancy \cite{KimHeungseob2017Rapc}. In the application of the active redundancy method, all modules would start operating simultaneously.  As they would need to be physically carried by robots this is likely to reduce power efficiency of robot operations. In cold standby, modules are protected from operational stress so that no redundant module can fail before it is used. 
It appears that the cold standby redundancy strategy is more suitable for SMR theory, which we apply here as our main tool to address reliability. There remain however modules, like inertial navigation modules, which may be required to be kept in active redundancy.

\subsection{Cost of maintenance}
In this section, the cost of maintaining is discussed to complete and update SMR theory, where maintenance tasks can be classified in three different ways: \emph{Preventive maintenance}, \emph{Corrective maintenance} and \emph{Failure-finding maintenance} as in system reliability theory \cite{rausand2003system}. 
\begin{enumerate}
   \item  \emph{Corrective maintenance (CM)} is executed after a module malfunctions.
   
  \item  \emph{Preventive maintenance (PM)} is a planned maintenance strategy when an item is activated and replaced regularly to prevent future failure.
 
  \item  \emph{Failure-finding maintenance (FFM)} is a special type of preventative maintenance that covers functional and operational diagnosis 
  to search for the next module to be replaced.
\end{enumerate}

PM and CM will be adopted for self maintaining robots to reduce the failure rate of whole system and extend survival time for a team of robots. The cost of maintenance is introduced to evaluate the cost of replacement of modules according to PM and CM respectively. FFM is outside the scope of this paper.

The SMR systems can mainly rely the corrective maintenance. It is effective where cost is more important then no interruption of work.   For robotic systems deployed in a dangerous situation such nuclear plants, the interruption of replacements, by gathering help from other robots, can potentially produce instability of operations, which can be more costly than robot maintenance.

\subsubsection{Corrective maintenance}
The purpose of corrective maintenance is to restore a module back to a functioning situation as soon as possible by substituting the failed module/sub-module by cold standby. This is also called breakdown maintenance or run-to-failure maintenance. 

For simplicity,  we assume that for a single module $i$, the probability of failure up to time $t$ follows an exponential probability distribution $ F_i(t)=1-e^{-\lambda_i t}$ and density $e^{-\lambda_i t}/\lambda_i$. Hence the mean time to failure (MTTF) for module $i$ is:
\begin{equation}
    MTTF(i)= \int_{0}^{\infty} te^{-\lambda_i t}/\lambda_i dt= \frac{1}{\lambda_i}
\end{equation}
The $\beta_i(t) = \lambda_i t$ is the mean value of the number of failures during an interval $t_0$. 
The  mean cost of maintaining module $i$ in a type $v$ robot by replacement over a time period $t_0$ is
\begin{equation}
    Ac_{iv}^{av}(t_0)=\frac {\beta^v_i(t_0) \gamma_{iv}}{t_0}= \lambda_i \gamma_{iv}
\end{equation}
where 
$\gamma_{iv}$ is the cost of one maintenance.  

The cost $Ac_{iv}$ can be applied in most cases of SMR applications, when the robotic system is working in time relaxed situation or the time of self-maintenance is negligible.

\subsubsection{Preventative maintenance}
Preventative maintenance aims to reduce the probability of failure of a module. Inspection, adjustments, lubrication, parts replacement, calibration and repair at times are applied to follow a maintenance policy. An active module of type $i$, or its parts, are replaced by cold-standby modules  periodically at times $t^i_0,2t^i_0,3t^i_0,...$. Planned cost and unplanned cost play a crucial role in the cost of maintenance. Compared with planned cost, unplanned cost is more disruptive and complicated, which is related with   factors such as problem investigation after robot malfunction and its rescue solution. 

\subsection{Quantification of reliability }  
System reliability with cold standby strategy, and perfect switching of modules, has been derived by Coit \cite{CoitDavidW2001Crof} for series-parallel systems. When applied to the $j$-th robot, the reliability, probability of non failure up to time $t$,  is
\begin{equation}
    R_j(t)=\prod_{i=1}^m (r_{i}(t)+\sum^{n_{i}-1}_{k=1} \int_{0}^{t} f^{k}_{i}(u)r_{i}(t-u)du)
    \label{eq:1}
\end{equation}
where $r_{i}(t)$ is distribution function of no breakdown happening to module $i$ up to time $t$ and $f_i(u)=1-r_i(t)$. $m$ is the number of module types, indexed by $i$ on each robot, and $n_i$ is the number of cold standby  $i$-modules available to the robot.
The above formula only applies to perfect switching and needs to modified to  

\begin{equation}
    R_{j}(t)=\prod_{i=1}^m (r_{i}(t)+p_i(t)\sum^{n_{i}-1}_{k=1} \int_{0}^{t} f^{k}_{i}(u)r_{i}(t-u)du)
    \label{eq:rel0}
\end{equation}
to take into account failure detection reliabilities.
Here $p_i(t)$ represents the detection reliability for module type $i$ on single robot, assuming detection is always made if a module fails.




A group of robots, as opposed to a single robot, has one great advantage when cold standby redundancy is applied: if they are structured similarly, they can share the same pool of modules to pick their replacements from.  The shared structure of each robot can then be optimised with consideration to the reliability of robots helping each other in module replacements when one breaks down. 

The group of robots do not need to be homogeneous as in practical applications various types of robots may need to work together. We assume in our reliability calculations that there are $V$ kinds of robots, with type indices $v=1,2,...,V$. In a team of robots, the number of type $v$ robots will be denoted by $v_l$.

\subsubsection{Full functional requirement}
For some robot deployments it can be a requirement that all members of a group of $v_l$ robots of type $v$ needs to remain operational to fulfil a mission. It means that  even if one robot fails, the whole robotic system would malfunction. Therefore, the system reliability of robots with type $v$ without cold-standby redundancy is
\begin{equation}
     R^v(t)= (R_j(t))^{v_{l}} 
    \label{eq:s1}
\end{equation}


\subsubsection{Minimal functional requirement}
 First consider the case when the robot group remains still usefully functioning until all robots fail. In this case, the failure of a robotic system can be defined by all robots  malfunctioning. The probability that at least one robot of type $v$ does not malfunction, out of $v_l$ of them, is:
\begin{equation}
    R_\mu^v(t)=1- (1-R_j(t))^{v_{l}} 
    \label{eq:p}    
\end{equation}
If the operational condition is that at least one robot needs to function from each type, then the probability of this happening is 
\begin{equation}
    R_\mu(t)= \prod_{v=1}^V [1- (1-R_j(t))^{v_{l}}] 
    \label{eq:p}    
\end{equation}

 \subsubsection{Partial functional requirement}
 A special case is when at least $m_l$ robots need to survive out of  $v_l$ type-$v$ robots in order that the whole group remains functionally useful. The probability of this can be calculated by 
\begin{equation}
    R_m(t)= \prod_{l=1}^V\sum_{k=m_l}^{v_l} \binom{v_l}{k} (R_j(t))^k(1-R_j(t))^{v_l-k} 
    \label{eq:s2}    
\end{equation}


\subsubsection{Reliability with cold standby redundancy}
Using modules kept on cold standby, a serial-parallel structure emerges, where the survival probability $r_i^v(t)$ can be updated to a higher probability. If $m^v_i$ modules are available from module type $i=1,2,...,m$ on cold standby, then the formula
\begin{equation}
    R_{j}^{v}(t)=\prod_{i=1}^{m^v} (r_{i}^v(t)+p^v_i(t)\sum^{m^v_i-1}_{k=1} \int_{0}^{t} f^{k}_{i}(u)r_{i}(t-u)du)
    \label{eq:c}
\end{equation}
is applicable to the requirement that robot $j$ remains functional up to time $t$, where  $p^v_i(t)$ is the probability of detection/switching success for module type $i$, which we also call the reliability of combined detection and switching. 
These two possibilities can however be separated as shown later.

Using the above formulae for each subgroups of robots of the same kind, the full functional reliability with spare modules can be worked out as:
\begin{equation}
    R(t)=\prod_{v=1}^{V} \prod_{j=1}^{v_l}R_{j}^{v}(t)
    \label{eq:all}
\end{equation} 
for the requirement that all robots need to remain fully functional. 


The probability for minimal reliability, which requires that at least one robot functionally survives from each type $v$,  is:
\begin{equation}
    R_\mu(t)=\prod_{v=1}^{V} (1-(1-R^v_j(t))^{v_{l}})
    \label{eq:ralls}
\end{equation}
The reliability of partial functional survival for each type of robot can be computed by products of (\ref{eq:s2}) for each robot type. 





Coit \cite{CoitDavidW2001Crof} presented reliability formula for cold-standby redundancy strategy with the use of exponential and Erlang distributions (cf. Ardakan \cite{ABOUEIARDAKAN2014107}). 
Underlying equations (\ref{eq:rel0})-(\ref{eq:ralls}) are homogeneous Poisson processes, hence, due to $ r_i(t)=  e^{-\lambda_i t}$, the $i$-th module's factors in (\ref{eq:c}) become
\begin{equation}
    R^v_{ji}(t)=e^{-\lambda^v_i t} +p^v_{i}(t)\sum_{k=1}^{n_{i}-1} \frac{e^{-\lambda^v_{i} t}(\lambda^v_{i} t)^k}{k! }  
    \label{eq:rel2}
\end{equation} 
where the $R^v_{ji}$ represents the reliability of module \textit{i} in robot \textit{j} of type  $v$. 

Based on (\ref{eq:rel2}), if $m_i$ denotes the total number of spare modules of type $i$ and $p_i(t)$ is the detection and replacement probability, then the {\it reliability of robot} $j$, up to time $t$, can be defined by 
\begin{equation}
    \tilde{R}^v_{j}(t)=\prod_{i=1}^m  R^v_{ji}(t)
    \label{exprel}
\end{equation} 

\begin{figure*}[h!]
\caption{ Tree-graph of self-maintenance. F1-F6 are activities for self-maintenance,  F7-F12 are capabilities needed and F13-F20 are modules supporting the capabilities. }
\includegraphics[scale=0.63]{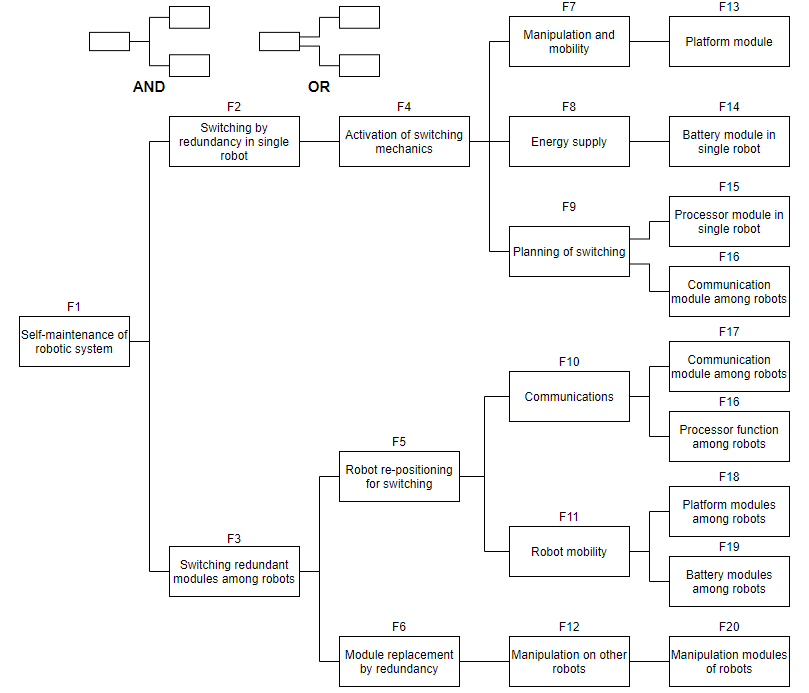} 
\label{fig:function}
\end{figure*}

\subsection{Types of module switching}
   A robot can possibly replace some of its modules by itself and some only by help of others. The reliability of the execution of switching module $i$ on its own, or by another robot, will be denoted by $p_i^s(t)$ and $p_i^o(t)$, respectively. There are a number of factors affecting switching reliability.
   A possible hierarchical decomposition of basic functionalities affecting switching are presented 
  in Fig. \ref{fig:function}. For an assessment some assumptions are made.
  
  \vspace{2mm}
  
\noindent \textbf{\emph{Assumptions: }} 
  \begin{itemize}
      \item Redundancy application is activated by cooperation of the platform module (locomotion module) and the processor module. 
      \item Switching reliability is defined by the reliability of switching to a redundant module. 
      \item Switching reliability is only dependent on the set 
      of active modules.  
       
  \end{itemize} 
  Self-maintenance can be split into  two main problems: switching redundancy on one robot and switching redundancy among a set of robots. Reliability of switching can be used to derive a solution for overall reliability assessment. Switching reliability on a single robot is primarily dependent on module redundancy on a single robot, while reliability of a team of robots depends on the overall availability of redundant modules in the team.  

The main modules, for which reliability is a key issue, are:
platform modules (locomotion modules), manipulator modules, battery modules, communication modules, processor modules and active self-protection modules (e.g. radiation protection modules). Their reliability functions will be respectively denoted by $r_{l}(t),r_{m}(t),r_{b}(t)$,$r_{c}(t),r_{p}(t),r_{a}(t) $.  Based on (\ref{exprel}), the reliability of a robot for its primary functions is 
\begin{equation}
\tilde{R}^v_{j}(t)\ =\ R^v_{jl}(t)R^v_{jm}(t)R^v_{jb}(t)R^v_{jc}(t)R^v_{jp}(t)R^v_{ja}(t)
\label{mainrel}
\end{equation}

  Communication activation includes the functioning of at least one of the processor and communication modules for diagnosis, reliability of the robot groups remaining capable of reporting about its conditions is
  \begin{equation}
      \tilde{R}^v_{jca}=  R^v_{jc}(t)R^v_{jp}(t)
  \end{equation}
  Reliability of a robot remaining able to move and manipulate is
   \begin{equation}
      \tilde{R}^v_{jlbp}=  R^v_{jl}(t)R^v_{jb}(t)R^v_{jp}(t)
  \end{equation}
  as the locomotion, battery and processor modules are needed for planning and for control of any motion execution. 
 Reliability of remaining manipulation capability can be obtained by  
  \begin{equation}
      \tilde{R}^v_{jbm}=  R^v_{jb}(t)R^v_{jm}(t)
  \end{equation}

\section{The Redundancy allocation problem}
In the previous sections, cold standby redundancy was used to enhance reliability. Other factors affecting the designer choice are price and quality of modules. The problem of optimising design of self maintaining robots has at least two main objectives: enhancing reliability and keeping running costs to a minimum. Reliability does however affect long term average maintenance costs through the need of replacements. This section presents the application of an evolutionary algorithm for the redundancy allocation problem. 

\subsection{Problem description} 

   The reliability of the group of $L$ robots of $V$ types is 

\begin{equation}
    \ R_\mu(t)=\prod_{v=1}^{V} [1-\prod_{j=1}^{v_{l}} (1- \tilde{R_{j}^{v}}(t)) ] 
    \label{eq:allrel}
\end{equation}
is to be maximised at a time $t$ of intended functional life of the robots. 

The total cost of the whole robotic system is
\begin{equation}
     \ C_\mu= \sum_{i}^{m} c_{i}m_{i}
\end{equation}
where $c_i$ is the cost of module $i$ and $m_{i}$ represents the quantity of modules $i$, including the active module and cold standby ones in the whole system.

As mentioned above, it may not be efficient for the robots to use a single repository of spare modules. If their working area spreads over large distances, then it is uneconomical and can cause long interruption of work for them to travel very far in order to pick up modules on cold standby. Spare modules are also impractical to be carried around by robots due energy consumption. 

If feasible, a midway solution is that each robot is allocated its own storage shed for spare parts, which is nearby to where the robot works. To be economic while providing the largest levels of reliability, some constraints 
\begin{equation}
      \ w_{ji} \leq w_{ji}^{limit} ,\ j \le L
\end{equation}
are set, where $w_{ji}$ is the number of modules of type $i$ for robot \textit{j} in its own storage and $w_{ji}^{limit}$ refers to the maximum  number of $i$ modules that a robot \textit{j} is allowed to keep in its storage.  

An alternative is to be non-specific about the upper limits for each robot and to use the general constraints 
\begin{equation}
      \ w_{ji} \leq w^{limit}_{i} ,\ j \le L,  i \le m
\end{equation}
Each robot may be defined a limited storage for spare modules, defined by:
\begin{equation}
      \ \sum_{i=1}^{m} w_{ji} \leq w^{limit}_{j} ,\ j \le L,  i \le m
      \label{eq:anothers}
\end{equation}
Note that the number of modules $i$ among all robots is
\begin{equation}
m_i= \sum_{j=1}^L w_{ji}
\end{equation}
and the distribution of $w_{ji}$ within $m_i$ does not affect the reliability formula  (\ref{eq:allrel}). If robots are limited to use their own storage, then their reliability needs to be modified to 
\begin{equation}
    R^w_{ji}(t)=r_{i}(t)+p^o_{i}(t)\sum_{k=1}^{w_{ji}-1} \frac{e^{-\lambda_{i} t}(\lambda_{i} t)^k}{k! }  
    \label{eq:relcrit}
\end{equation} 
which then provides a more specific 
\begin{equation}
   \ R^w_{all}(t)=\prod_{j=1}^{L} [1-\prod_{i=1}^{w_{ji}} (1-R^w_{ji}(t))] 
    \label{eq:allrel1}
\end{equation}
that can be a much reduced reliability level but with reduced costs of switching to spare modules. 

In any design optimisation the choice of constraints and application of reliability functions is dependent on the engineering and commercial constraints persisting. 


\subsection{Complexity of designing for reliability}
Without using the local storage upper limit quotas $w_{ji}^{max}$,  the number of evaluations for cost-reliability pairs is 
\begin{equation}
    X= \prod_{i=1}^m m^{max}_i
    \label{eq:hirel}
\end{equation}
where $ m^{max}_i$ is the maximum practical number possible for each module type $i$. Alternatively, with the use $w^{max}_{ji}$, the complexity is
\begin{equation}
    X^w = \prod_{i=1}^m   \prod_{j=1}^L w_{ji}^{max}
    \label{eq:lowrel}
\end{equation}
which can be significantly higher than (\ref{eq:hirel}).

In a practical implementation of self-maintaining robots working in an isolated area, it is expected that spare modules are regularly provided to the robots by their human supervisors. If the supervisors maintain the optimised redundancy level, then the robots will keep up the associated reliability level. Under such assumptions, reliability and cost of modules can be jointly applied in the computation of the time averaged continuous running cost of the robot team, which then becomes 
\begin{equation}
    C_{cont} = L\sum_{i=1}^m c_i/\lambda_i
\end{equation}
In many future applications the robots are deployed for a fixed period with the number of cold standby modules to be determined. In these cases the optimisation problem becomes a multi-objective one, balancing reliability against overall costs. The complexity of this problem is NP-hard in terms of the number of robots as the variations of cases to be evaluated are as in (\ref{eq:hirel}) or (\ref{eq:lowrel}) .


\subsection{Alternative optimisation methods}

The NP-hard optimisation for self-maintenance can be solved for a low number of robots 
by discrete evaluation of all cases. For large number of robots and modules, however,  alternative optimisation methods are needed. As an example,  next  we present the application of an evolutionary optimisation algorithm. 

\subsection{Use of evolutionary optimisation}

Deb \cite{DebK2002Afae} developed a non-dominated sorting-based multi-objective  EA called NSGA-II. Their aim was to replace the original NSGA algorithm, which had a number of shortcomings such as computational complexity and lack of elitism. We found NSGA-II particularly effective for optimisation that involves two objectives.

In NSGA-II, the population is initialized first. The population is ranked, depending on non-domination into fronts 1, 2, 3,... and so on. The first front is a set with completely non-domination individuals, while the second one is dominated by the individuals in the first front only, and this domination relationship carries in recursively. Each front is assigned a fitness measure to rank all individuals, which is called the 'crowding distance'. The crowding distance is used to evaluate  distances between individual and their neighbours. A larger average crowding distance can represent a better diversity in the population. A binary tournament selection process is run, which is based on the rank and crowding distance to pick up parents. An individual can be selected for its lower rank than others for its crowding distance being better than that of others. Finally, the offspring and current population is sorted again by dependence on non-domination and a population of fixed size \textit{N} is selected for the next generation. The algorithm fits well the two-objectives problem, which strikes a 'trade-off' mechanism between cost and system reliability.          
This is the main reason of using  NGSA-II in this example, with its 'trade-off' property, preferred over the MOEA/D \cite{QingfuZhang2007MAME} with weighting vectors.         

\subsection{An Example}

Assume that a robotic system has 6 robots of types $v \in \{1,2,...,6\}$ and that  each individual has 6 modules with redundancy limits  $w^{max}_j=6$. We also assume that only one robot is used from each type, so that $v_l=1, l=1,...,6$. 
The replaceable core modules are: the platform module, battery module, processor module, manipulator modules, telecommunication module and active radiation prevention module.  Each module has three quality types with varying system reliability and costs. 
A simple exponential function is used for the distribution of failure times, and known $\lambda$ rates of failure are used to define the system reliability functions for each module.

\subsubsection{Chromosome representation}
\begin{figure*}[h]
\centering
\caption{Chromosome representation depending on an example}
\includegraphics[scale=0.55]{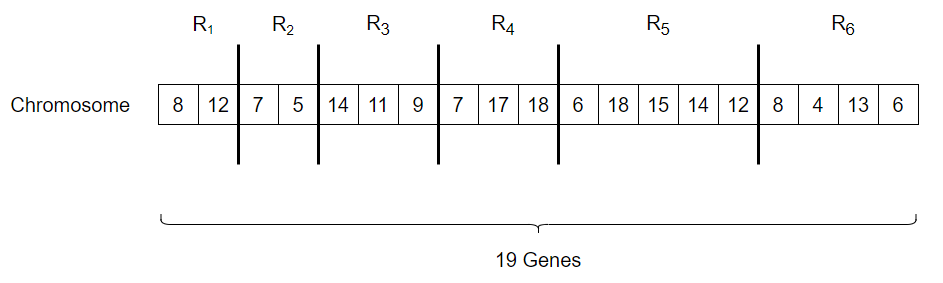}
\label{fig:chromosome}
\end{figure*}

The genes of a chromosomes represent the module's type by an integer, so the length of  the chromosome is the number total spare connectors in a robotic configuration before redundancy allocation as in Fig. \ref{fig:chromosome} . In the example, there are 18 modules, so 
the number of spare connectors is 19, the length of the chromosome. 

\begin{table*}
\centering
    \addtolength{\leftskip} {-1.5cm} 
    \addtolength{\rightskip}{-1.5cm}
\caption{Data used in the modules}
\label{tab:my-table}
\resizebox{\textwidth}{!}{
\begin{tabular}{|l|c|c|c|c|c|c|c|c|c|}
\hline

\multicolumn{1}{|c|}{$m_i$} & 1      & 2      & 3      & 4      & 5      & 6      & 7      & 8      & 9      \\ \hline
\multicolumn{1}{|c|}{cost ($c_i$)}                & 2000   & 2300   & 2400   & 200    & 230    & 280    & 400    & 420    & 450    \\ \hline
\multicolumn{1}{|c|}{$\lambda_i$(months)}                          & 0.0031 & 0.0032 & 0.0034 & 0.0050 & 0.0052 & 0.0057 & 0.0034 & 0.0036 & 0.0037 \\ \hline
\multicolumn{1}{|c|}{$m_i$}                       & 10     & 11     & 12     & 13     & 14     & 15     & 16     & 17     & 18     \\ \hline
\multicolumn{1}{|c|}{cost ($c_i$)}                    & 300    & 310    & 360    & 1600   & 1700   & 1900   & 800    & 820    & 870    \\ \hline
\multicolumn{1}{|c|}{$\lambda_i$(months)}                           & 0.0021 & 0.0022 & 0.0023 & 0.0012 & 0.0013 & 0.0016 & 0.0076 & 0.0077 & 0.0079 \\ \hline
\end{tabular}}
\end{table*}

\begin{table}

\centering
\caption{Robot Configuration without redundancy}
\label{tab:my-table}
\scalebox{1.3}{
\begin{tabular}{|c|c|l|l|c|c|}

\hline
\multicolumn{1}{|l|}{Robot number} & \multicolumn{3}{l|}{Total slot} & \multicolumn{1}{l|}{Free Slot} & \multicolumn{1}{l|}{Configuration Code} \\ \hline
1                                  & \multicolumn{3}{c|}{6}          & 2                              & 1,4,7,10                                \\ \hline
2                                  & \multicolumn{3}{c|}{6}          & 2                              & 1,3,8,16                                \\ \hline
3                                  & \multicolumn{3}{c|}{6}          & 3                              & 1,5,8                                   \\ \hline
4                                  & \multicolumn{3}{c|}{6}          & 3                              & 1,4,7                                   \\ \hline
5                                  & \multicolumn{3}{c|}{6}          & 5                              & 2                                       \\ \hline
6                                  & \multicolumn{3}{c|}{6}          & 4                              & 3,9                                     \\ \hline
\end{tabular}}
\end{table}

\begin{figure*}[h]
\centering
\caption{Multi-optimization figures of test problem using NGSA-II}
\includegraphics[scale=0.48]{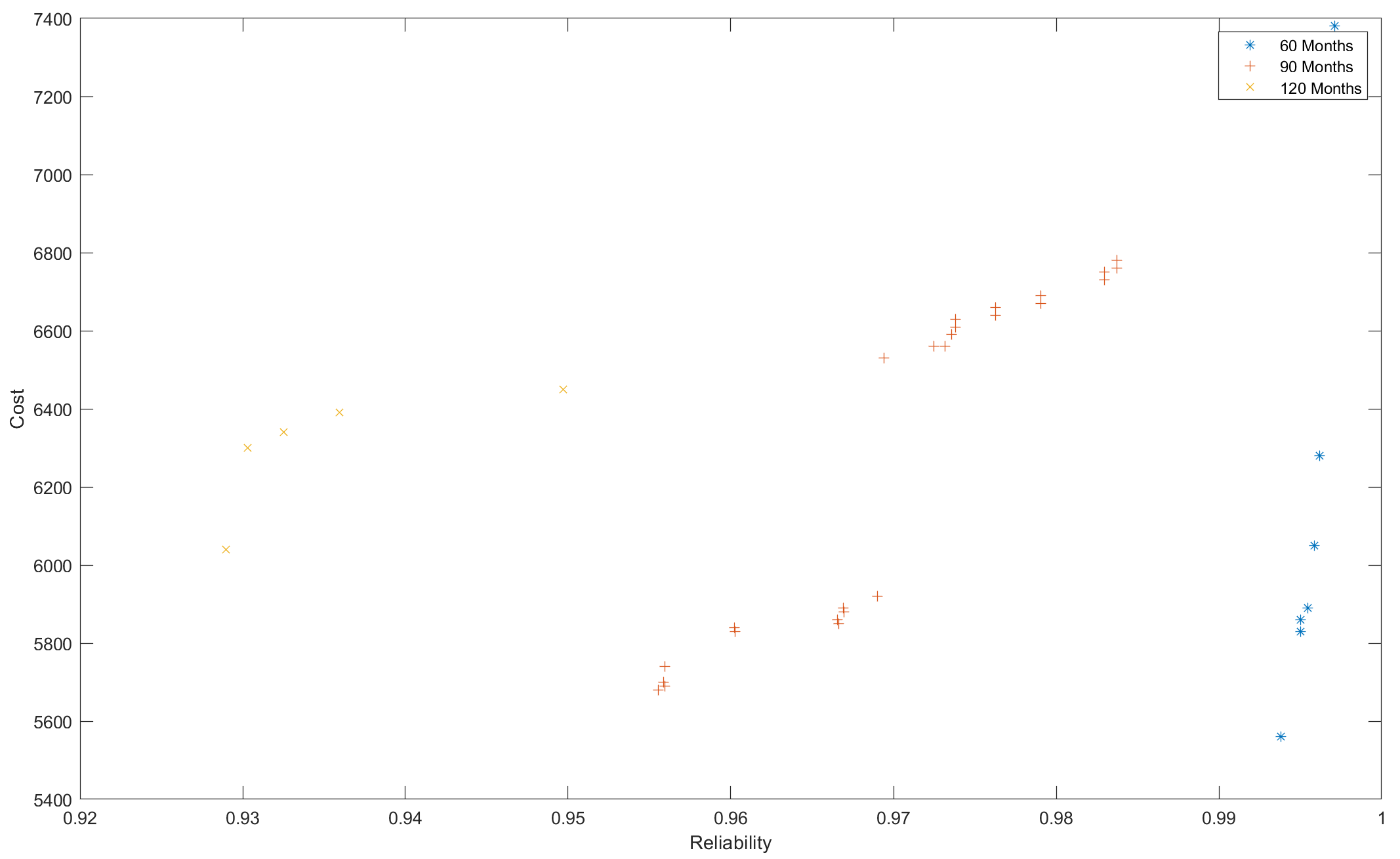} 
\label{fig:ngsa}
\end{figure*}

\subsubsection{Demonstration of results}
Based on the assumptions above, a group of results have been calculated in Matlab as in Fig. \ref{fig:ngsa}. The \textit{X} coordinate displays the system reliability beyond a time $t$, when the \textit{Y} represents cost of the whole system. The points (star, plus and cross) in Fig. \ref{fig:ngsa} show the non-dominated area in three different working times (60,90,120 months).  In Fig. \ref{fig:ngsa} , the blue stars illustrate the system reliability of a robotic system in 60 months, which can maintain a very good reliability close to  1 and with a low cost configuration. Then, as the working time is increasing (90 months and 120 months), though the system reliability of the robotic system is weakening, it is still higher than the 0.9, which means that the whole system is working well for 120 months after missions started. It is illustrated here that with  a cold-standby strategy the robotic system always maintains a reasonably good working state for a long term periods. 

\section{Principles of connector mechanisms for SMR systems}
Connectors or docking mechanisms are an integral part of any long lasting robotic system. They allow the attachment of different modules, components  and sometimes also robots.  The purpose of a connector or docking mechanism is to transfer mechanical forces and moments, electrical power and communication data in a robotic system. 
Reliable and automatic connection between modules is widely recognized as a critical design element of any such system. In this section we introduce the main principles of connector and docking mechanism designs. 

\subsection{Related work}
There are many published solutions to connectors that can be relevant to the theory of self-maintaining robots.  Yim \cite{YimM2000Pamr} \cite{YimM2002Cadf} designed a connector for Polybot, which is a T-shaped hook with four bolts on  a cross shaped spring loaded blade. It does not however lends itself to automated  connection with ease.
Jorgensen \cite{JorgensenM.W2004MAmf} built a hook based bonding mechanism for the Atron modular robotic system. The shape of connector is positioned on a sphere, making it difficult to provide a stable and strong connection.
Hossain \cite{Connector} designed a rotary-plate genderless single sided docking mechanism, which performs robustly and efficiently in unstructured terrains. The connector is deployed in ModRED robots \cite{c4}.
It is clear that most of the existing modular robotic systems utilize some mechanical methods, including hooks and bolts based solutions to form a tight connection between two modules. 
Murata\cite{MurataS2002Msmr} proposed a permanent magnetic docking mechanism to connect M-TRAN modules. A spring system had also been developed to be able to break the connection between modules. Each module has three male and three female permanent magnets.  In addition to the permanent magnetic solution, White \cite{WhiteP.J2004Sscr} put forward the idea of using electro-magnetic connectors. 
Neubert \cite{NeubertJonas2014SCfM} developed a strong, lightweight, and solid-state connection approach depending on heating a low melting temperature alloy to form reversible soldered connections without any moving component. The connector is appropriate for autonomous actions, which can connect and disconnect with a target module without external manipulation. But the connectors can also be deployed in dangerous environments such as active  or decommissioned nuclear plants as the melting temperature is higher than for solder alloys.

\subsection{Design principles for SMR systems}
There are a series of requirements for self maintaining robots with regards to docking mechanisms, some of which had been adopted  from the theory of modular and self-reconfigurable robots. 

\begin{itemize}
\item \emph{Fast and easy}:
The connector structure must be easy and quick to  lock and unlock by manipulator modules. 
\item \emph{Compact}:
Each connectors must be compact in size and weight, which enables the attachment and use of multiple connectors. 
\item \emph{Strength}:
The connection should be strong and robust to sustain mechanical loads. 
\item \emph{Fault tolerance}:
Establishing a connection should be robust under various  environmental disturbances at the time of locking. 
\item \emph{Unilateral actuation}:
Connection should be made possible even if  one of the modules is functionally unresponsive, i.e. when at least one side of the connectors is live. 
\item \emph{Long life and adaptability}:

The connector should carry on functioning for long time periods under various environmental conditions. 
\item \emph{Standardization (consistency)}:
Maximum compatibility of connectors are to be provided across all modules to increase ability to diagnose faults and sustain maintainability. 
\item \emph{Diversity}:
For self maintaining robots there are some key benefits in supporting multiple types of connectors in one module or one system to increase the efficiency and functionality of the whole robotic system, especially for self-maintaining ability. It increases reliability if a module  can also be connected  with more than one type of connectors.
\item \emph{Reversibility and repeatability}:
Both forming and breaking of connections must be smoothly and reliably  operating over long  periods of operations.  
\end{itemize}

\subsection{Types of connectors for SMR systems}
Two types of $A$ and $B$ connectors are proposed for connections between core modules and payload modules. So far we have only dealt with modules, which maintain the survival capabilities of robots. In practice there can be modules on a robot, which purely serve a mission capability, such as sensing and special tools for manipulation of materials. These  modules will be called {\it payload modules} as they only support the profitability of a robot mission. 
\subsubsection{Connector A}
Connector type $A$ is proposed to be used to establish and break connections between a core modules and a payload modules. As such (1) their size and weight need to be carefully optimised; (2) their wired information transmission capability to avoid electromagnetic interference; (3) their  ability to transmit electrical power; (4) their ability to transfer mechanical force to keep payload modules securely attached and (5) they need to be as inexpensive as possible due to their relatively large numbers needed in long missions.

\subsubsection{Connector B}

Connector type $A$ is proposed to be used to establish and break connections between core modules. These connectors also need their size, weight optimised; capable of carrying internal communication channels, which are vital to revive failed robot basic functionality; capable to transfer electrical power specifically between any two core types of core modules; capable of carrying mechanical force to keep core modules in correct place.
For example, one robot stuck in the soft terrain and isn't able to move out, then a rescue robot can connect with the robot stuck using connector B to sustain the mechanical load needed for rescue. 


\section*{Conclusions}

In this paper we have developed some of the theoretical foundations of self-maintaining robots with the aim of assisting future robotic developments for applications where autonomous or remotely supervised robots need to work on their own, without physical contact with humans. The best candidates for replaceable modules and components have been discussed. System reliability has been addressed from the viewpoints of structural reliability and functional reliability. Itself the reliability of detection has been accounted for. Maintenance types have been identified as preventative (PM), corrective(CM) and fault finding(FFM) and their cost functions have been established. 
Formulae have been provided for reliability over finite time horizon for minimal and partial functional requirements. Computations for reliability under cold standby of components and modules has been presented, inclusive replacement/switching reliability for teams of robots with homogeneous and heterogeneous architectures. Computations have been provided for long term operational capability of a team of robots over infinite time horizons.  Complexity issues of design optimisation of self-maintaining robots have been addressed and an evolutionary computation example provided. One of the most important electro-mechanical components, universal connectors for both mechanical strength and electrical reliability, have been illustrated. The basic theory presented can be refined  further in individual designs of future robots in the nuclear, space,  nature preservation  areas, and also in dangerous environments of industrial laboratories. 
 


\bibliographystyle{unsrt}
\bibliography{bibtex.bib}

\end{document}